\documentclass[runningheads]{llncs}
\usepackage{amsmath}
\usepackage{graphicx}
\usepackage{multirow}

\begin{document}
\title{User profiles matching for different social networks based on faces identification}
\titlerunning{User profiles matching in social networks}
\author{Timur Sokhin\inst{1} \and
Nikolay Butakov\inst{1} \and
Denis Nasonov\inst{1}}
\authorrunning{T. Sokhin et al.}
\institute{ITMO University, 49 Kronverksky Pr., St. Petersburg, 197101, Russia
\email{245591@niuitmo.ru}\\ 
\email{alipoov.nb@gmail.com}\\
\email{denis.nasonov@gmail.com}\\ }

\maketitle
\begin{abstract}It is common practice nowadays to use multiple social networks for different social roles. Although this, these networks assume differences in content type, communications and style of speech. If we intend to understand human behaviour as a key-feature for recommender systems, banking risk assessments or sociological researches, this is better to achieve using a combination of the data from different social media. In this paper, we propose a new approach for user profiles matching across social media based on publicly available users' face photos and conduct an experimental study of its efficiency. Our approach is stable to changes in content and style for certain social media
\keywords{face detection, profiles, matching, social networks, face embedding, clustering, computer vision}
\end{abstract}
\section{Introduction}
Nowadays, social media may differ in their capabilities to share information and express, that reflects in types of published content, conversation style, etc. For instance, About.me or LinkedIn may be used as the main page for self-presentation purposes, while Twitter or Instagram are used for informal communication of random people, publishing selfies.

Understanding how a user behaves is an important task for many applications such as the generation of recommendations, candidate assessment by HR departments or even for the analysis of further developing for social media itself. We suppose that a person comprehensively may be described with a set of profiles joined from different social networks. While some users link all their profiles together explicitly or mention in their posts somehow, mostly, people don't want to associate them.

Previous attempts to solve this problem has been directed to matching by features such as names, friend-graphs, published textual contents (e.g. topics of posts) and so on. These methods often may lack precision or recall because of differences between social networks in published content and style, absence of required interlinks for friends and so on, and lead to mismatching of expected and real person \cite{DBLP:journals/corr/ZhongCKLS17}. 

In this work, we propose and study a new approach of profiles matching based on publicly available users' images and faces identification. The face is a unique attribute for humans, that should keep almost unchanged from network to network. The existing methods of face detection and embedding allow us to detect faces on photos and compare them. But a single face image may suffer from positions, perspective, quality problem. We need more than that to reliably match profiles: we have to identify the owner's faces among others, even if there is only one person presence on a photo. 

The contributions of this paper are the following: (1) we propose a novel approach to user profiles matching using face detection and comparison of face embeddings from different social media; (2) we conduct a set of experiments for two popular in Russia social networks VKontakte and Instagram and investigate limitations of our approach in terms of quality and quantity of a data. The latter includes answering the following questions:

\begin{enumerate}
    \item How many data (photos) does effective matching require?
    \item How does efficiency (precision and recall) depend on the quality and the quantity of the data?
\end{enumerate}
\section{Related work}
Mostly, previous work in this field is focused on the easily accessible information about the user: self-description, biography, name, nickname \cite{Goga2014MatchingUA} \cite{Malhotra2012}; or on the dynamic of users behaviour: dates of posts, profiles updates \cite{Zhang2015} \cite{Hazimeh17}. As it noticed in \cite{Shu2015} and \cite{Khaled2017}, this kind of information (username, location,  followers/followings, meta paths) are very noisy, easily faked, not required, they provide huge research of existing methods to profiles matching. The last suppose that methods of behaviour dynamic analysis show potential for further work, but they have some major disadvantages: they require collecting of information during some period of user activities and require an unusual method of data representation in different social media, which can vary in their features.

Also, it should be noticed, that there is almost no works with images, which provide a lot of additional information about the user itself and are useful for profiles matching. \cite{Zhang2015}, \cite{Hazimeh17} and similar approaches require features, which can not be extracted from all social media - Instagram and VKontakte are different in the type and the context of the content, friends-system, etc. Our approach reveals new possibilities for comparing profiles based on photographic materials, which are more suitable in this case.
\section{The Approach}
The main idea of our approach is to form a single defining vector - representation of a user's profile based on the embeddings of his faces.
\paragraph{Data Collecting.}
Our approach consists of several stages. At first, we must data from two social media using a crawling framework (profiles, photos from albums and posts) \cite{butakov2018unified}. For the purposes of validation of our results, we collect a set of profiles from VKontakte, which have an explicit link to their secondary profile in Instagram - the only possible way to build the labelled dataset.
\paragraph{Face Detection and Embedding.}
We process photos using two algorithms:
\begin{enumerate}
    \item face detection - we apply MTCNN - Multi-task Cascaded Convolutional Networks \cite{zhang2016joint}, which achieved efficiency superior to the closest competitors and is not affected by scaling of the faces;
    \item face embedding - to construct embeddings of extracted faces FaceNet neural network is applied \cite{schroff2015facenet}.
\end{enumerate}
We apply MTCNN pre-trained on the WIDER FACE dataset and FaceNet pre-trained on the VGGFace2\footnote{Code repository used - https://github.com/davidsandberg/facenet}. Then this data is filtered.
\paragraph{Filtering.}
The extracted face embeddings are further filtered by their parameters according to several heuristics: 
\begin{enumerate}
    \item filtering by number of pixels (hereinafter, we will use the term quality of the image);
    \item filtering by anchors (child faces removing).
\end{enumerate}

FaceNet has limitations on the minimum required quality of images and we filter images of faces by the number of pixels of these faces. The accurate control of the above parameters allows to achieve an improved precision and recall of matching, this is partly due to the behaviour of the selected method for embedding construction. In the experimental study in Sect. \ref{exp} we found an effect of the quality of facial images on the final matching efficiency - it improves the F1-score by 4\%.

\begin{figure}[h]
  \centerline{\includegraphics[scale=0.5]{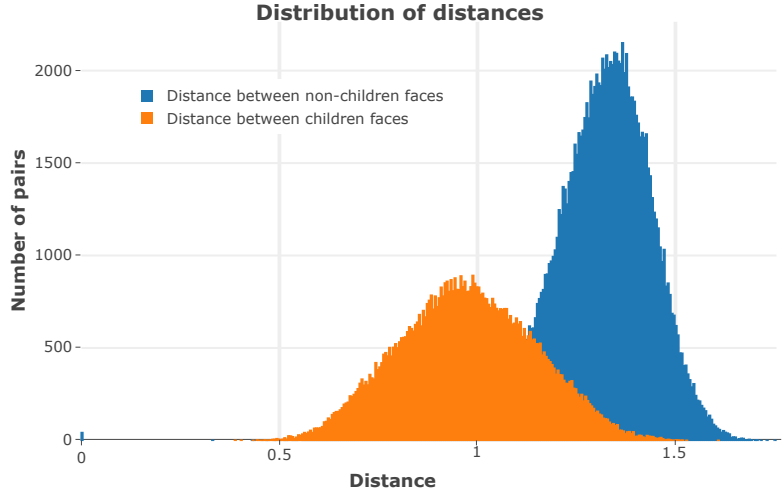}}
  \caption{Distribution of distances between random people faces and between children faces}
  \label{fig:distance}
\end{figure}

The other heuristics probably can be related to the dataset limitation of VGGFace2 with which FaceNet was trained. VGGFace2 contains young and mature faces of people but does not contain the faces of babies and small children. This leads to a problem that embeddings of child's faces have a very small margin between each other. That is why we should remove their faces from the user's collection of photos to avoid mismatching of profiles. Figure \ref{fig:distance} reveals that the distribution of distances between embeddings of children's faces has a bias from the distribution of distances between embeddings of random people's faces.

Additional filtering of data is accomplished using so-called anchors. An anchor is a vector that represents some space of embedded faces. In our study, we use the anchor to represent the faces of children. We create it by following way. A set of children faces was collected semi-automatically: we find kindergarten and photographers accounts using tags and specific usernames. For instance, tags under the photos with words "children", "kindergarten", etc. Then we build an anchor - element-wise mean of all vectors of children's faces. All face embeddings which are close to this anchor are removed from the dataset.
\paragraph{Owner identification.}\label{owner_detection}
This is the main part of our approach that is performed separately for each profile in each social network. Embeddings of faces are formed in Euclidean space. We apply hierarchical clustering for each profile separately with the single linkage algorithm and distance threshold 0.8. This algorithm allows us to generate a non-fixed number of clusters based on the Euclidean distance between face embeddings.

Each cluster of the profile should belong either to a single person in the real world, whose faces have slightly different but close embeddings or to persons who look very similar due to distortions introduced by hairstyle, put on glasses, beards and other things which make them look similar.
This is the main part of our approach that is performed separately for each profile in each social network. Embeddings of faces are formed in Euclidean space. We apply hierarchical clustering for each profile separately with the single linkage algorithm and distance threshold 0.8. This algorithm allows us to generate a non-fixed number of clusters based on the Euclidean distance between facial embeddings.

Each cluster of the profile should belong either to a single person in the real world, whose faces have slightly different but close embeddings or to persons who look very similar due to distortions introduced by hairstyle, put on glasses, beards and other things which make them look similar.

We assume that most users publish photos with different people, but the number of their face occurrences is greater than others. Following this hypothesis, in order to find the owners' faces, we must choose the largest cluster and combine them into one vector - the defining vector (DV) of profile using faces from a chosen cluster. The DV is an element-wise mean of all generated embeddings with the same dimension (\ref{DV}, where V - face embedding, n - number of embeddings of the user).

\begin{equation}\label{DV}
    DV = \frac{1}{n}\sum_{i=1}^{n}V_{i}
\end{equation}

\begin{figure}
  \centerline{\includegraphics[natwidth=843,natheight=459,scale=0.20]{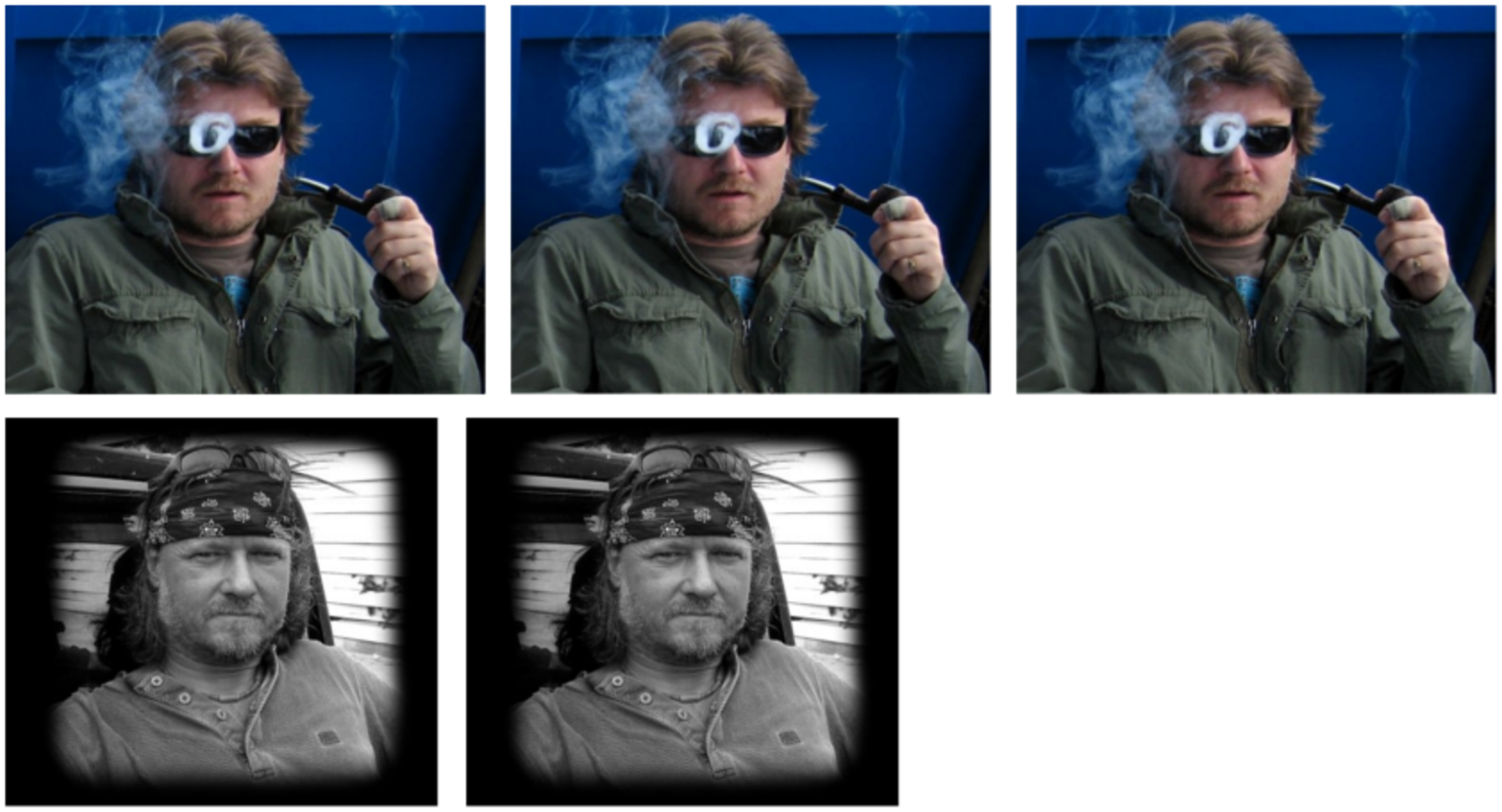}\includegraphics[natwidth=843,natheight=459,scale=0.20]{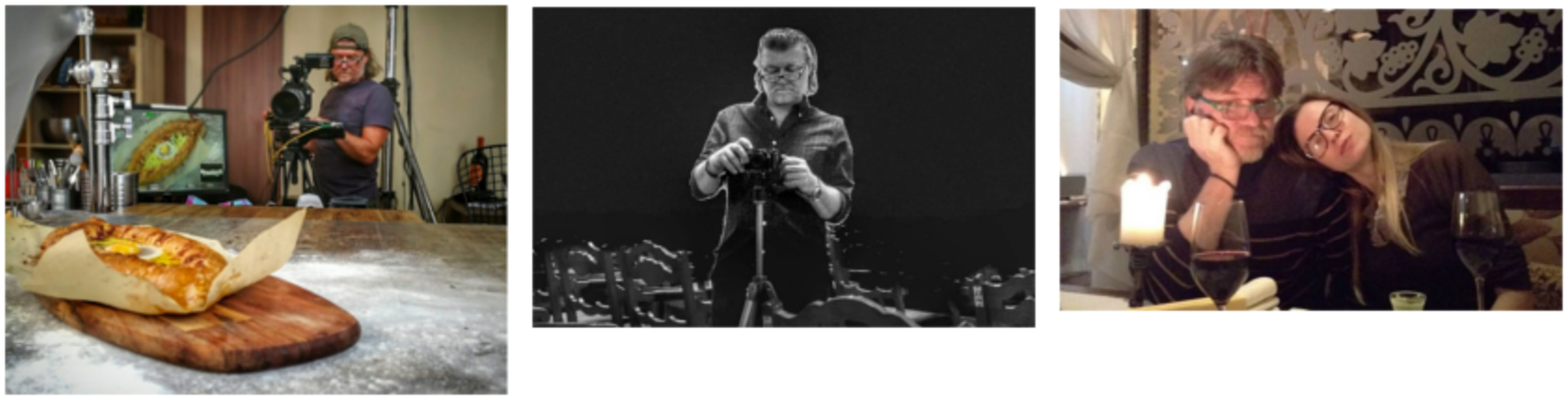}}
  \caption{Examples of cluster: (a) the first largest; (b) the second largest}
  \label{fig:sharp}
\end{figure}

However, due to possible sharpness of the DV, it is worth to take into account the other largest clusters. Sometimes people publish many similar photos, even the same photos. In case that is shown in Fig. \ref{fig:sharp} (a) the first cluster only consists of two unique images. We are not able to match this profile using this cluster. But we can add the others (for instance, the second largest, that is shown in Fig. \ref{fig:sharp} (b)) and form a new DV using more than two unique face embeddings. Our experiments in Sect. \ref{exp} show that this assumption and the proposed solution allow us to achieve results that exceed the use of one cluster. Experimental results give us the optimal value - 2 clusters. If after clustering there is only one cluster, we use all photos of the user, if there are all clusters with the same size (e.g. 1 element), we set this profile as "unable to set the owner" and mark as profiles without a pair.

After that, the DV of each profile in both social media represents the user and will be used to matching. If the size of the largest cluster is less than a given threshold, this user marked as profiles without pair, because it is not possible to detect the owner's face correctly. The tuning of the threshold is also provided in the experimental study.
\paragraph{Profiles Matching.}
The process of profiles matching is simple: defining vectors of users from two social media are compared with each other. We calculate the L2 norm between profiles in two social media, for each profile in one social media we find the profile from the other with the smallest distance and mark as a candidate for matching (\ref{argmin}).
\begin{equation}\label{argmin}
\begin{split}
    argmin \; L2\left(DV_{i}^{VK}, DV_{j}^{Inst}\right) = \left\{\right. DV_{j}^{Inst} | DV_{k}^{Inst} \in DV^{Inst} : \\
     L2\left(DV_{i}^{VK}, DV_{k}^{Inst}\right) > L2\left(DV_{i}^{VK}, DV_{j}^{Inst}\right)\left. \right\}
\end{split}
\end{equation}

If the smallest distance is higher than the given threshold (threshold distance, hereinafter), this means there is no pair in the other social media or we could not find it.
\section{Experimental study}\label{exp}

\subsection{Details of the experimental part}
\subsubsection{Our experimental plan} consists of three main steps: baseline evaluation using real names-based matching; evaluation for full profiles without any limitations; evaluation with alignment rate reduction and photos number reduction.

\subsubsection{Dataset description.} 
We use our own dataset - Dataset4675, which consists of 4675 profiles from VKontakte and 3100 profiles from Instagram, which simulates working with partially aligned networks - only 3100 VKontakte users have a pair in other social media. Dataset4675 users have from 50 to 500 publicly available photos.

\subsubsection{Metrics.}
We clarify definitions of precision, recall and F1-score, that we use for this classification problem, which is not fully classical. Since we are working with VKontakte as our main social media and want to saturate its profiles with additional information, all metrics are calculated with respect to the number of VKontakte users.

With V as a number of all real pairs in our dataset (3193), K\textsuperscript{p} as a number of the correct predictions of the algorithm (correctly matched pairs of VKontakte and Instagram profiles) and K as a number of all predictions of the algorithm, the precision is defined as follows (\ref{P}):
\begin{equation}\label{P}
    P = \frac{K^p}{K}
\end{equation}
And the recall is defined as follows (\ref{R}):
\begin{equation}\label{R}
    R = \frac{K^p}{V}
\end{equation}
We need both the recall and precision in order to evaluate our approach, F1-score shows the balance between them and is used to choose the best parameters.

\subsection{Baseline evaluation. Real names matching}
The real names of users from Dataset4675 are compared with Levenshtein distance metric and sensitivity is analyzed according to its threshold distance. For each user we are looking for the closest user from other social networks, if the closest distance exceeds the threshold value, we remain this user without a pair. The real names are processed in the following sequence: lower case translation; non-alphabetic characters removing; transliteration.

The precision and recall are shown in Table~\ref{table:RN}. The highest F1 of 0.295 is achieved with P=0.765 and R=0.183 and the distance threshold of 4 permutations. With a small dataset in relation to the real number of users, this approach achieves a good precision, but it should be noticed that the precision decreases with the increasing number of users. This can be explained from the fact of a large number of homonyms in the real world. Also, we have a very low recall rate.

\begin{table}
\caption{Real name based matching results}\label{table:RN}
\centering
\begin{tabular}{llll}
\hline\noalign{\smallskip}
Threshold & Precision & Recall & F1-score\\
\noalign{\smallskip}
\hline
\noalign{\smallskip}
1 & 0.976 & 0.106 & 0.191 \\
2 & 0.972 & 0.148 & 0.257 \\
3 & 0.922 & 0.169 & 0.286 \\
{\bfseries 4} & {\bfseries 0.765} & {\bfseries 0.183} & {\bfseries 0.295} \\
5 & 0.511 & 0.192 & 0.279 \\
6 & 0.352 & 0.198 & 0.253 \\
7 & 0.269 & 0.203 & 0.231 \\
8 & 0.235 & 0.205 & 0.219 \\
\hline
\end{tabular}
\end{table}
\subsection{Evaluation for full profiles}
\subsubsection{Cluster analysis.}At first, we analyze the dependency on the clusters number in Table~\ref{table:Clust} with fixed parameter of threshold distance - 0.65 and image quality - 6400. 
\begin{table}
\caption{Cluster dependence analysis}\label{table:Clust}
\centering
\begin{tabular}{llll}
\hline\noalign{\smallskip}
Number of largest clusters used & Precision & Recall & F1-score\\
\noalign{\smallskip}
\hline
\noalign{\smallskip}
1 & 0.9617 & 0.7885 & 0.8665 \\
2 & {\bfseries 0.9782} & {\bfseries 0.7875} & {\bfseries 0.8726} \\
3 & 0.9797 & 0.7839 & 0.8709 \\
4 & 0.9793 & 0.7845 & 0.8712 \\
5 & 0.9801 & 0.7842 & 0.8713 \\
\hline
\end{tabular}
\end{table}

It can be seen as proof of the requirements of more than 1 clusters mentioned in Sect. \ref{owner_detection} - the F1-score in this case is 0.855. The optimal value of the number of the cluster is 2.
\subsubsection{Face-based matching.}
\begin{figure}[h]
   \centerline{\includegraphics[scale=0.35]{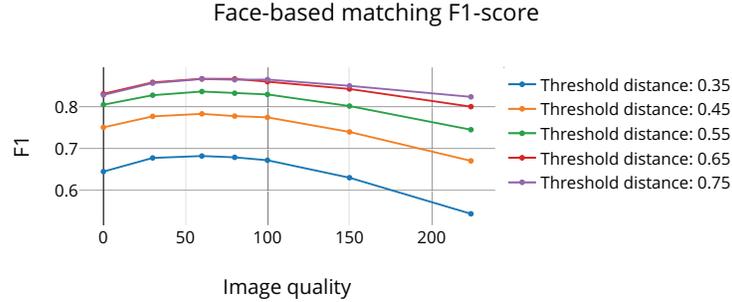}}
  \caption{F1-score of face-based matching depending on the image quality and the threshold distance}
  \label{fig:face_all}
\end{figure}

We also provide a sensitivity analysis of our approach in Fig.~\ref{fig:face_all}. We use Dataset4675 for this part of the experimental study. One can see a strong dependence between the threshold distance and efficiency. While a high precision is achieved with a smallest threshold distance value, the recall remains lower than 0.7, that can be seen in Table~\ref{table:FaceAll}. The higher F1 is 0.0868 with image quality 80 and threshold distance 0.65.

\begin{table}[!htbp]
\setlength{\tabcolsep}{12pt}
\caption{Face-based matching results}\label{table:FaceAll}
\centering
\begin{tabular}{llllll}
\hline\noalign{\smallskip}
\multirow{1}{*}{} & \multicolumn{5}{c}{Threshold distance} \\

Image quality                  & 0.35   & 0.45  & 0.55  & 0.65  & 0.75  \\
\noalign{\smallskip}
\hline
\noalign{\smallskip}
\multicolumn{6}{c}{Precision}                              \\
\noalign{\smallskip}
\hline
\noalign{\smallskip}
0 & 0.997   & 0.989   & 0.976   & 0.951   & 0.898   \\
30 & 1.0   & 0.999   & 0.997   & 0.984   & 0.933   \\
60 & 1.0   & 1.0   & 1.0   & 0.995   & 0.947   \\
80 & 1.0   & 1.0   & 1.0   & {\bfseries 0.994}   & 0.946   \\
100 & 1.0   & 1.0   & 1.0   & 0.992   & 0.948   \\
150 & 1.0   & 1.0   & 1.0   & 0.992   & 0.948   \\
\noalign{\smallskip}
\hline
\noalign{\smallskip}
\multicolumn{6}{c}{Recall}                                 \\
\noalign{\smallskip}
\hline
\noalign{\smallskip}
0 & 0.478   & 0.606   & 0.687   & 0.739   & 0.77   \\
30 & 0.513   & 0.637   & 0.709   & 0.763   & 0.793   \\
60 & 0.519   & 0.645   & 0.721   & 0.77   & 0.8   \\
80 & 0.515   & 0.638   & 0.715   & {\bfseries 0.77}   & 0.798   \\
100 & 0.507   & 0.634   & 0.71   & 0.761   & 0.797   \\
150 & 0.461   & 0.588   & 0.671   & 0.734   & 0.772   \\
\hline
\end{tabular}
\end{table}
\subsection{Evaluation with the reduced alignment rate and the reduced number of photos}
Here we experiment with limited data and rate of alignment of users. If our approach requires as much data as possible, it is only applicable for government and law enforcement with social media cooperation.
\subsubsection{Avatars only matching.} 
\begin{figure}[h]
  \centerline{\includegraphics[scale=0.35]{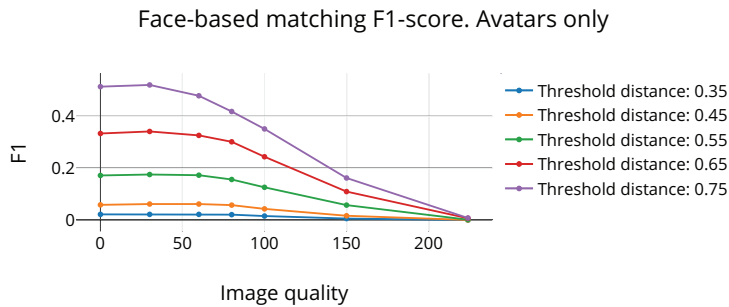}}
  \caption{F1-score of face-based matching depending on the image quality and the threshold distance. Avatars only}
  \label{fig:face_avatars}
\end{figure}

When working with facial images, using avatars can be the easiest way. This removes the need for the owner detection stage because the idea of an avatar is to present the owner. Here we use only users' avatars from Dataset4675 to evaluate this assumption in Fig.~\ref{fig:face_avatars}.

We faced the recall decrease in general and almost zero F1-score with a high value of the quality filter. We achieve 0.539 F1-score with the following parameters: threshold distance - 0.75, quality - 30.
\subsubsection{Reducing the number of images for each user.}
\begin{figure}[h]
  \centerline{\includegraphics[scale=0.35]{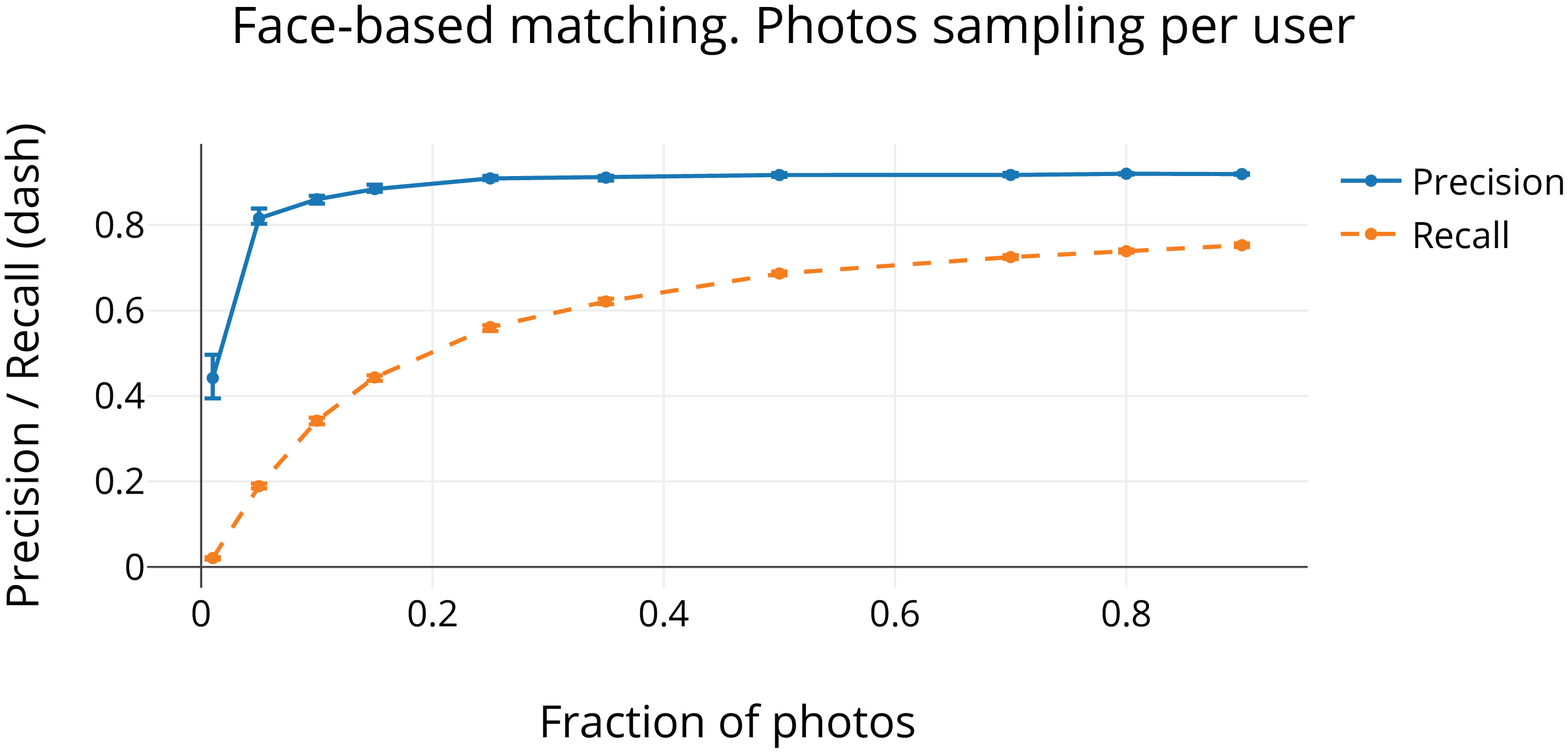}}
  \caption{The dependence of the efficiency of the algorithm on the proportion of user photos}
  \label{fig:sampling}
\end{figure}

We reduce the number of available photos of each user from Dataset4675 in order to estimate our approach in the condition of greater uncertainty in Fig.~\ref{fig:sampling}. 

The procedure of sampling is as follows: for each user, we select X\% of his/her photos for 10 times. It is interesting that the precision rate remains almost the same even with 10\% of data from each user profile of both social media. The reason for the low recall rate is the owner detection part: a small amount of randomly sampled data does not allow to find the owner's face and to form a good defining vector.

\subsubsection{Reducing the rate of intersections. Partial alignment.}
\begin{figure}[h]
  \centerline{\includegraphics[scale=0.35]{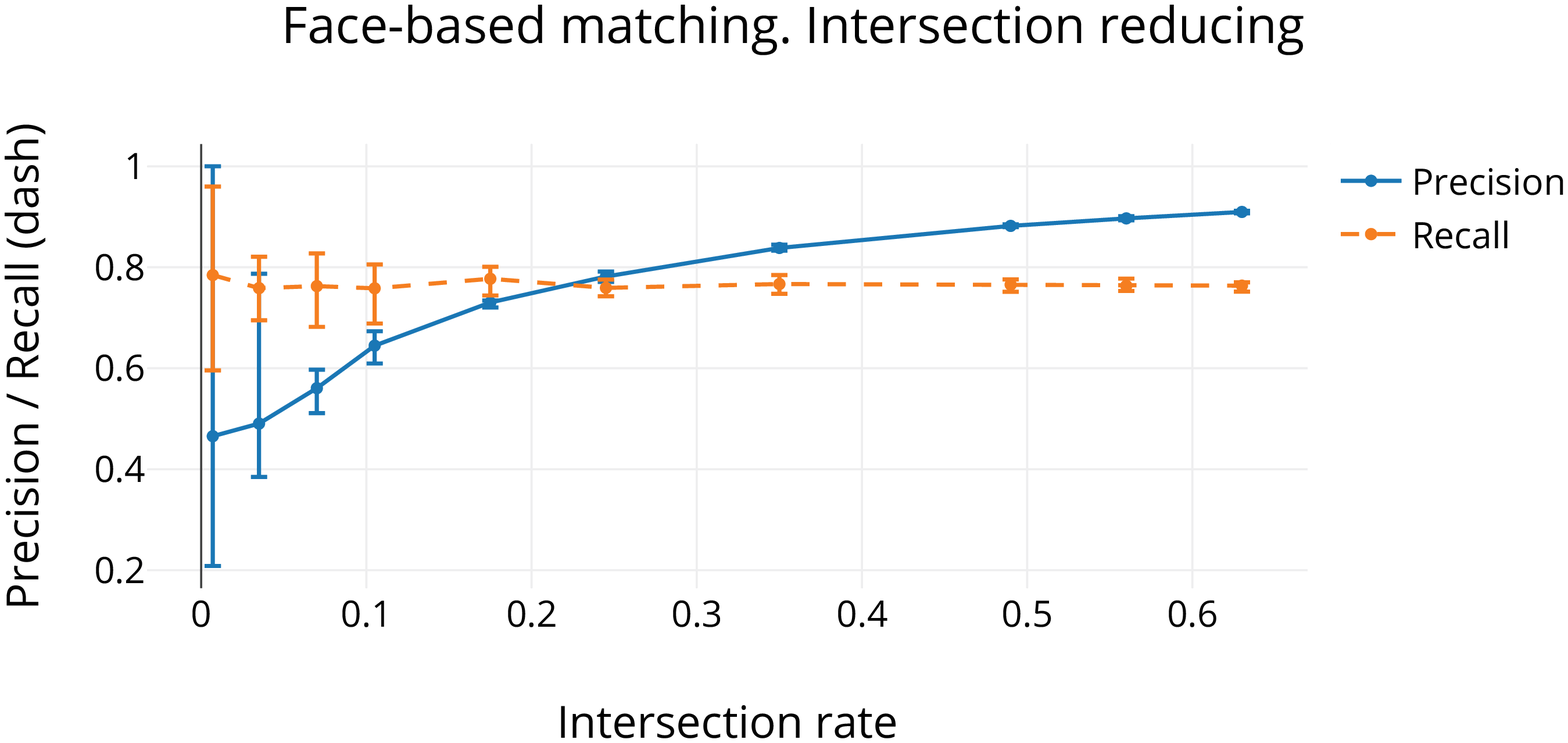}}
  \caption{The dependence of the efficiency of the algorithm on the proportion of user photos}
  \label{fig:inter}
\end{figure}

In the final part of the experiments, we examine the partial alignment of social networks. As noted by \cite{Zhang2015} authors real social media are partially alignment - not all users from one social media have accounts in another one. It is impossible to investigate the real rate of this intersection, but we can consider a number of rate values and create a synthetically reduced intersection. The high variance of precision and recall depicted in Fig.~\ref{fig:inter} is explained by user properties: we match different users, due to random sampling. Some of these users could have more or fewer photos, good or bad (such as biased vector) defining vectors. 
The stability of recall shows that our approach can be applied on low-alignment networks. The precision decreased on low-rate alignment because of many false-positive samples, this can potentially be improved by additional filtering.
\section{Discussion}
The results of faces-based profiles matching with only avatars show a low efficiency - the recall is 0.375 and the precision is 0.963), which is due to the following:
\begin{itemize}
    \item the quality of user avatars are not always enough, this leads to unnecessary filtering and decreasing of recall value, there was only 57\% of faces from avatars with quality over 80;
    \item as shown in \cite{DBLP:journals/corr/ZhongCKLS17}, almost 25\% of Facebook users have two people on the avatar - we cannot detect the owner using this kind of images, and the defining vector is not precise.
\end{itemize}

There is also one indirect reason why avatars are not enough even if we were able to detect the owner: as it is seen in Table~\ref{table:Clust}, one cluster gives us less F1-score - 0.8665. This aspect and the analysis of results show that very homogeneous clusters lead to mistakes in matching. Using only one image would be a degenerate case of one cluster from one face.

The results of our study indicate that our approach works less efficiently without all available user's data. This is expected behaviour, because of the essence of our approach: we work with a content of profiles. The recall decreases very quickly, but the precision remains almost the same until the 5-10\% of available data (P=0.80, R=0.18 with 5\% of available photos and P=0.84, R=0.33). But even 18\% of users still allow you to match many profiles in absolute values.

The last thing to discuss the experiments is user sampling. It should be noticed: we do not know the real intersections of people in different social media. \cite{VK-stat} and \cite{Inst-stat} reports that there are 30 millions of Instagram users in Russia and 80 millions of VKontakte users. Also, we know that 3.3 millions of VKontakte users link their Instagram profile. So, the rough estimate of profiles alignment is 3-4\%. This value allows us to achieve P=0.49 and the average R=0.758. The alignment rate is probably greater due to historical features: VKontakte is one of the first social media in Russia and it is very popular among active users of the Internet who can be Instagram users. In this case, the alignment is about 30\% and the expected precision is 0.8 and the recall is 0.76.
\section{Conclusion}
In this paper, we propose a method to profiles matching across different social media using users' photos. Our approach use photos from the profiles to form a single feature-vector using embedding techniques and use only this vector for further profiles matching. The proposed approach achieved a high precision up to 0.994 in case of 70\% of users have profiles in both social media and recall up to 0.76. Profiles can be matched even with the limitations described in the experimental section.

Our approach provides a large number of applications. We can match a set of criminals faces from street or security cameras with their profiles in social media. Moreover, it is very useful for scientific purposes: additional information could help to find new features of the user behaviour and open new opportunities in the research of social media impact on the person.
\bibliographystyle{splncs03}
\bibliography{body}

\end{document}